\def\year{2020}\relax
\documentclass[letterpaper]{article} 
\usepackage{aaai20}  
\usepackage{times}  
\usepackage{helvet} 
\usepackage{courier}  
\usepackage[hyphens]{url}  
\usepackage{graphicx} 
\usepackage{graphicx}  
\usepackage[linesnumbered,ruled]{algorithm2e} 
\usepackage{multirow}
\title{Heuristic Black-box Adversarial Attacks on Video Recognition Models}
\author{
Zhipeng Wei\textsuperscript{\rm 1}, 
Jingjing Chen\textsuperscript{\rm 2}, 
Xingxing Wei\thanks{Corresponding author}\textsuperscript{\rm 3}, 
Linxi Jiang\textsuperscript{\rm 2},
Tat-Seng Chua\textsuperscript{\rm 4},  \\
\Large \textbf{Fengfeng Zhou\textsuperscript{\rm 5},
Yu-Gang Jiang\textsuperscript{\rm 2}}\\
\textsuperscript{\rm 1}Jilin University,
\textsuperscript{\rm 2}Fudan University,
\textsuperscript{\rm 3}Beihang University,
\textsuperscript{\rm 4}National University of Singapore,\\
\textsuperscript{\rm 5}Health Informatics Lab, College of Computer Science and Technology, and Key Laboratory of Symbolic \\ Computation and Knowledge Engineering of Ministry of Education, Jilin University, Changchun, Jilin, China, 130012\\
weizp17@mails.jlu.edu.cn, chenjingjing@fudan.edu.cn, xxwei@buaa.edu.cn, lxjiang18@fudan.edu.cn,\\ chuats@comp.nus.edu.sg, FengfengZhou@gmail.com, ygj@fudan.edu.cn}
 
\begin{document}

\maketitle

\begin{abstract}
We study the problem of attacking video recognition models in the black-box setting, where the model information is unknown and the adversary can only make queries to detect the predicted top-1 class and its probability. Compared with the black-box attack on images, attacking videos is more challenging as the computation cost for searching the adversarial perturbations on a video is much higher due to its high dimensionality. To overcome this challenge, we propose a heuristic black-box attack model that generates adversarial perturbations only on the selected frames and regions. More specifically, a heuristic-based algorithm is proposed to measure the importance of each frame in the video towards generating the adversarial examples. Based on the frames' importance, the proposed algorithm heuristically searches a subset of frames where the generated adversarial example has strong adversarial attack ability while keeps the perturbations  lower than the given bound. Besides, to further boost the attack efficiency, we propose to generate the perturbations only on the salient regions of the selected frames. In this way, the generated perturbations are sparse in both temporal and spatial domains. Experimental results of attacking two mainstream video recognition methods on the UCF-101 dataset and the HMDB-51 dataset demonstrate that the proposed heuristic black-box adversarial attack method can significantly reduce the computation cost and lead to more than 28\% reduction in query numbers for the untargeted attack on both datasets.
\end{abstract}

\section{Introduction}
Deep neural networks are vulnerable to adversarial samples  \cite{goodfellow2014explaining,szegedy2013intriguing}. 
Recent works have shown that adding a small human-imperceptible perturbation to a clean sample can fool the deep models, leading them to make wrong predictions with high confidence \cite{moosavi2017universal}. As results, it has raised serious security concerns for the deployment of deep models in security-critical applications, such as face recognition \cite{kurakin2016adversarial}, video surveillance \cite{sultani2018real},
etc. Therefore, it is crucial to study the adversarial examples for deep neural networks. In addition, investigating adversarial samples also helps to understand the working mechanism of deep models and provides opportunities to improve networks' robustness. 

Nevertheless, most of the existing works focus on exploring adversarial samples for image recognition models under white-box or black-box settings \cite{kurakin2016adversarial,szegedy2013intriguing,papernot2016limitations,carlini2017towards,su2019one}. Adversarial attacks on video models, especially the black-box attacks, have been seldom explored. Indeed, deep neural network based real-time video classification systems, e.g., video surveillance systems, are being increasing deploys in real-world scenarios. Therefore, it is crucial to investigating the adversarial samples for video models.

This paper studies the problem of generating adversarial samples to attack video recognition models in the black-box settings, where the model is not revealed and we can only make queries to detect the predicted top-1 class and its probability. The major obstacle for solving this problem is how to reduce the number of queries to improve attack efficiency. Compare to statistic images, the dimensionality of video data is much higher as videos have both temporal and spatial dimensions. As results, directly extending the existing black-box attack methods proposed for image models to video models will consume much more queries for gradient estimation, leading to the low attack efficiency. 

\begin{figure*}[t]
  \centering
  \includegraphics[width=0.8\textwidth]{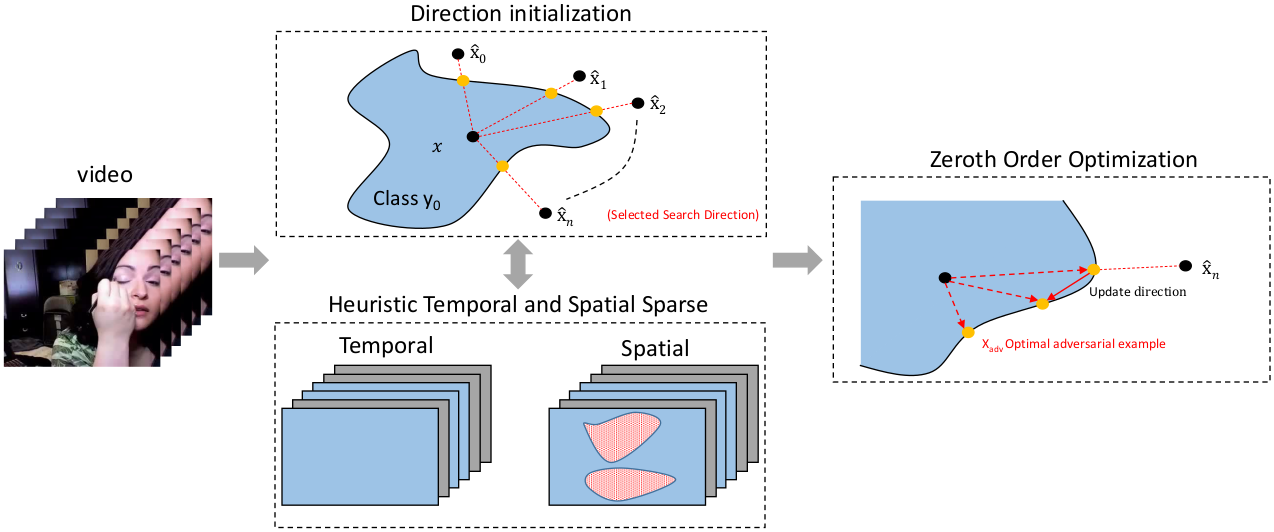}
  \caption{Overview of the proposed heuristic algorithm for black-box adversarial attacks on video recognition models.}
  \label{fig:over}
\end{figure*}

To this end, we propose an efficient black-box attack method specially for video models in this paper. Considering the fact that video data contains both temporal and spatial redundancy, it is therefore not necessary to search the adversarial perturbations for all the frames \cite{Wei2019SparseAP}. Besides, for a given video, different frames and different regions in each frame contribute differently towards video classification results \cite{peng2018two}. Intuitively, the key frames which contain the key evidence of one specific event, or the salient regions in the key frame (usually the foreground regions), play a vital role for the video classification results. As results, generating adversarial perturbations on the salient regions of key frames is the most effective way to attack video models. Motivated by this, we propose a heuristic-based method to select the key frames for adversarial samples generation. Figure \ref{fig:over} overviews the proposed method. In the proposed method, the perturbation directions are initialized with the samples from other classes different from the class of the clean video. Meanwhile, it heuristically searches a subset of frames from the clean video with the initialized directions, and the salient regions detected from the selected frames are utilized to update the initial directions. The initial directions as well as the selected key frames are updated iteratively to make the number of selected frames as small as possible while keeps the perturbation noise lower than the given bound. Then the initial direction which has the minimum distance to the decision boundary is chosen as the final initial  direction and zeroth order optimization \cite{chen2017zoo} is next utilized for direction updating. As the perturbations are generated only for the salient regions of the selected frames, the proposed method 
significantly reduces the number of queries. 
Our major contributions can be summarized as follows:
\begin{itemize}
\item We propose an efficient black-box adversarial attack model that heuristically select a subset consisting of the  key frames to generate adversarial perturbations.
\item We introduce saliency detection in the black-box adversarial attack model to generate adversarial perturbations only on the salience regions of the key frames, which further reduces 
query numbers. As far as we know, this is the first attempt to introduce prior knowledge for black-box video attack.
\item Extensive experiments on two benchmark data sets demonstrate that the proposed method is efficient and effective. It reduces more than 28\% reduction in query numbers for the untargeted attack. 
\end{itemize}

\section{Related Work}
\label{related_work}
\subsection{Adversarial Attack on Image Models}
Recent works on adversarial attack are mostly focus on image models, including both white-box \cite{szegedy2013intriguing,goodfellow2014explaining,kurakin2016adversarial,dong2018boosting,carlini2017towards,moosavi2016deepfool,papernot2016limitations,sarkar2017upset} and black-box attacks \cite{chen2017zoo,tu2018autozoom,cheng2018query,brendel2017decision,papernot2017practical,liu2016delving}. This section reviews recent works on generating adversarial samples for image models. 
\subsubsection{White-box Attack}
White-box attack assumes that the structure as well as the parameters of the targeted model are known to the attacker. In this setting, attacks can be done easily as the gradient of attack objective function can be computed via backpropagation. In recent years, different white box attack methods have been proposed. For example, L-BFGS attack \cite{szegedy2013intriguing} first crafts adversarial examples against deep neural networks towards maximizing the network’s prediction error. and demonstrates the transferability of adversarial examples. Fast Gradient Sign Method (FGSM) \cite{goodfellow2014explaining} updates gradient once along the direction of the sign of gradient at each pixel. FGSM can be applied in multiple times with small step size for better adversarial examples in \cite{kurakin2016adversarial},  and also be combined with momentum to increase adversarial intensity \cite{dong2018boosting}. C\&W's Attack \cite{carlini2017towards}  generates the targeted adversarial examples by defining effective objective functions, and provides three types of attacks: $l_0$ attack, $l_2$ attack, and $l_\infty$ attack. Other algorithms such as DeepFool \cite{moosavi2016deepfool}, Jacobian-based Saliency Map Attack (JSMA) \cite{papernot2016limitations}, UPSET and ANGRI \cite{sarkar2017upset}, etc, have carried out white box attacks from different aspects.

\subsubsection{Black-box Attack}
Compare with the white-box attack, the black-box attack is a more realistic but more challenging setting as the model information remains unknown to the attacker. Existing efforts on black-box attack for image models include Zeroth Order Optimization (ZOO) \cite{chen2017zoo}, Autoencoder-based Zeroth Order Optimization Method (AutoZOOM) \cite{tu2018autozoom}, decision based black-box attack \cite{brendel2017decision}, Opt-attack \cite{cheng2018query}. Zeroth Order Optimization (ZOO) \cite{chen2017zoo} uses ADAM's update rule or Newton's method to update perturbations with the formulation of the C\&W attack, gradients estimated by symmetric difference quotient, and coordinate-wise Hessian. Autoencoder-based Zeroth Order Optimization Method (AutoZOOM) \cite{tu2018autozoom} is a query-efficient black-box attack, which employs an autoencoder in reducing query counts, and the random full gradient estimation. Decision-based black-box attack \cite{brendel2017decision} starts with an adversarial perturbation and performs a random walk on the boundary to reduce the perturbation. Opt-attack \cite{cheng2018query} formulates the black-box attack into a real-valued optimization problem, which requires fewer queries but finds smaller perturbations of adversarial examples. There also exist transfer-based attacks using a white-box attack towards the substitute model for generating adversarial examples in a black-box setting, such as \cite{papernot2017practical}, \cite{liu2016delving}.

\subsection{Adversarial Attack on Video Models}
There is much less related work on generating adversarial samples to attack video models compared with adversarial attacks on image models. The first white-box attack method for video recognition models is proposed in \cite{Wei2019SparseAP}, where an $l_{2,1}$-norm regularization based optimization algorithm is proposed to compute the sparse adversarial perturbations for videos. Different from \cite{Wei2019SparseAP}, \cite{Li2018StealthyAP} took advantages of Generative Adversarial Networks (GANs) to do a white-box attack. Their attack method generates the universal perturbation offline and works with unseen input for the real-time video recognition model. Nevertheless, the above mentioned methods assume the complete knowledge of the video recognition models which is different from our settings. Another work is \cite{inkawhich2019adversarial} which proposes an untargeted adversarial attack that based on FGSM \cite{Goodfellow2015ExplainingAH} and iterative FGSM \cite{Kurakin2017AdversarialML} for flow-based video recognition models in both white-box and black-box settings. The proposed white-box attack leverages Flownet2 and chain rule to obtain the gradients through the whole video recognition model, where Flownet2 is required to estimate optical flow frames while also providing gradient information easily. They used the transferability of adversarial examples generated by the white-box model to implement the black-box attack. 
Different to these two works which generate the adversarial perturbations for all the frames in the video, our work heuristically searches a subset of frames and the adversarial perturbations are only generated on the salient regions of the selected frames, which effectively reduce 
the number of queries. 

\section{Methodology}
\label{methodology}
In this section, we introduce the proposed heuristic black-box attack algorithm for video recognition models. We assume that the predicted top-1 class and its probability are known to an attacker. 

Denote the video recognition model as a function $F$. Specifically, the DNN $F(x)$ takes a clean video $x \in R^{T \times W \times H \times C}$ as an input and output the top-1 class $\widehat{y}$ and its probability $P(\widehat{y}|x)$, where $T$, $W$, $H$, $C$ denote the number of frames, width, height, and the number of channels respectively. It suffices to denote its associated true class label by $y \in Y = \{1, \cdots, K\}$, where K is the number of classes, and its adversarial example by $x_{adv}$. In the untargeted attack, the goal is to make $F(x_{adv}) \neq y$, while for the targeted attack with the targeted adversarial class $y_{adv}$, the goal is to satisfy $F(x_{adv}) = y_{adv}$.

Our attack algorithm is built based on the Opt-attack \cite{cheng2018query}, which is originally proposed for attacking image models by formulating the hard-label black-box attack as a real-valued optimization problem that can be solved by zeroth order optimization algorithms. We extend Opt-attack \cite{cheng2018query} from image models to video models. Specifically, to deal with the high dimensional video data and improve the attack efficiency, we propose a heuristic algorithm to select a subset of video frames and only search the adversarial perturbations for the selected frames. Following the Opt-attack, we denote $\theta$ as the search direction of the video. The function of the distance from $x$ to the decision boundary along the direction $\theta$ is denoted as $g(\theta)$, which is calculated by a fine-grained search and a binary search function defined in \cite{cheng2018query}. The objective of Opt-attack's is to find the direction that minimizes $g(\theta)$ by zeroth order optimization proposed in \cite{cheng2018query}. In Opt-attack algorithm, the optimization process can be divided into two parts: direction initialization and direction updating. The direction initialization is done by the function $\theta = \frac{p}{\Arrowvert p \Arrowvert}$, where $p = \hat{x}-x$, and $\hat{x}$ denotes the video from other classes; Then update $\theta$ with the zeroth order optimization and $g(\theta)$. Finally, we find the adversarial example by $x_{adv} = x + g(\theta^*) \times \theta^*$, where $\theta^*$ is the optimal solution of the zeroth order optimization. 

Here, we focus on the direction initialization part, and propose heuristic temporal and spatial selection methods in direction initialization for generating barely noticeable adversarial videos that result in misclassification of DNN $F$. Specifically, we attempt to find the sparse $\theta$ by modifying $p$ with the equation $p = p \times M$, where $M \in \{0,1\}^{T \times W \times H \times C}$ is the mask introducing temporal and spatial sparsity. For each element in $M$, if its value equals 1, the corresponding pixel in the video will be included in the process of perturbation computation. The selected regions with $M$ are named as masked regions. During the adversarial example generation, the computed perturbations are only added to the masked regions. Next, we describe in detail the proposed heuristic sparse methods.

It worth to mention that it has been demonstrated in \citeauthor{uesato2018adversarial}, the type of gradient-free optimizer has relatively small influence towards the attack efficiency. Hence, we use the zero-order optimization here. 
The estimated gradient is defined as 
\begin{equation}
    \hat{g} = \frac{g(\theta + \beta \textbf{u}) - g(\theta)}{\beta }\cdot \textbf{u}
    \label{estimate_gradient}
\end{equation}
where \textbf{u} is a random Gaussian vector of the same size as $\theta$, and $\beta > 0$ is a smoothing parameter which will reduce by 10 times if the estimated gradients can't provide useful information for updating $\theta$. Following \cite{cheng2018query}, we set $\beta = 0.005$ in all experiments. Besides, we sample \textbf{u} from Gaussian distribution for 20 times to calculate their estimators and average them to get more stable $\hat{g}$. In each iteration, $\theta$ is updated by
\begin{equation}
    \theta \gets \theta - \eta \hat{g}
    \label{update_theta}
\end{equation}
where $\eta$ is the step size, which is adjusted at each iteration by a backtracking line-search approach.   

\subsection{Heuristic Temporal Sparsity}
\label{temporal}
Videos have successive frames in the temporal domain, thus, we consider to search a subset of frames that contributes the most to the success of an adversarial attack. Here we introduce the concept of temporal sparsity which refers to the adversarial perturbations generated only on the selected key frames during the direction initialization process. 

In order to achieve temporal sparsity, we first propose a heuristic-based algorithm to evaluate the importance of each frame. Given a frame $t$, $M_{t}$ means that all frames except the t-th frame are equal to 1. Based on $F$, we can get top-1 class label and its probability by $\widehat{y_t}, P(\widehat{y_t}|(p\times M_t+x)) = F(p\times M_t+x), \forall t \in \{1,\cdots ,T\}$. We then sort the sequence of video frames according to the descending order of $P(\widehat{y_t}|(p\times M_t+x))$ under the condition that $\widehat{y_t}$ is an adversarial label. Note that, the larger value of $P(\widehat{y_t}|(p\times M_t+x))$ means the less importance of t-th frame towards generating the adversarial sample.

We search a set of key frames based on the sorted video frame sequence. This searching process is performed during the direction initialization for all randomly selected $\hat{x}$. Denote $\omega$ as the bound of mean absolute perturbation of each pixel in videos that determines when to stop toward searching a smaller set of key frames. When the mean absolute perturbation on the selected key frames lower than the bound $\omega$, our method will continue searching a smaller set of key frames for adversarial perturbation generation. The value of $\omega$ is selected according to the experiments on validation set, which we will discuss in the experimental part. Algorithm \ref{key_frame} summarizes the whole procedure of searching the key frames for the targeted attack. In Algorithm \ref{key_frame}, \textbf{DELFRAME} sets the values of $i$-th frame in $M$ to $0$; \textbf{SORTED} sorts the indexes of frames in descending order by the associated top-1 probabilities; \textbf{LENS} calculates the number of selected key frames; and \textbf{MAP} computes the mean absolute perturbation for $g(\theta) \times \theta$.


\begin{algorithm}[t]
    \label{key_frame}
    \SetKwInOut{Input}{Input}
    \SetKwInOut{Parameter}{Parameter}
    \SetKwInOut{Output}{Output}

    \Input{DNN $F$, clean video $x$, true label $y$, target class $y_{adv}$, initial mask $M \in \{ 1 \}^{T \times W \times H \times C}$,  an empty array $A$.}
    \Output{Mask of key frames $M$.}
    \Parameter{Bound $\omega$.}
    $\hat{x} \gets$ a video sample of target class $y_{adv}$;
    
    $p,k \gets \hat{x} - x, 0$ \;
    \For{$ t\gets 1$ \KwTo $T$}
    {$M_t \gets \mathrm{DELFRAME}(M, t)$ 
    \tcp*{the values of $i$-th frame are equal to 0.}
    $\widehat{y}, P(\widehat{y}|(p \times M_t + x)) \gets F(p \times M_t +x)$ \;
    \uIf{$\widehat{y} = y_{adv}$}{$A[k],k \gets (t, P(\widehat{y}|(p \times M_t+x))), k+1$\;}{}
    }
    $A \gets \mathrm{SORTED}(A)$
    \tcp*{indexes of frames are sorted in descending order by $P(\widehat{y}|(p \times M_t+x))$.}
    $\theta_{init} \gets \frac{p}{\Arrowvert p \Arrowvert}$
    \For{$i \gets 1$ \KwTo $k$}
    {
    $\widehat{M} \gets\mathrm{DELFRAME}(M, A[i])$ \; 
    $\widehat{p} \gets p \times \widehat{M}$ \;
    $\theta \gets \frac{\widehat{p}}{\Arrowvert \widehat{p} \Arrowvert}$\;
    $\widehat{y}, P(\widehat{y}|(x+\widehat{p})) \gets F(x + \widehat{p})$ \;
    
    \uIf{$\widehat{y} = y_{adv}$}
    { 
    
    \eIf(){$\mathrm{MAP}(g(\theta) \times \theta) \leq \omega$} {
        \uIf(\tcp*[f]{the number of key frames.}){$\mathrm{LENS}(\widehat{M}) < \mathrm{LENS}(M)$} {$M, \theta_{init} \gets \widehat{M}, \theta$\;} {}} {\uIf{$\mathrm{MAP}(g(\theta) \times \theta)<\mathrm{MAP}(g(\theta_{init}) \times \theta_{init})$} {$M, \theta_{init} \gets \widehat{M}, \theta$\;}{}} 
    
    }{}
    
    }
    \Return $M$
    \caption{Heuristic temporal selection algorithm for the targeted attack.}
\end{algorithm}

\subsection{Heuristic Spatial Sparsity}
Intuitively, salient regions, for example, the foreground of the frames, contribute more to the video classification results. Generating adversarial perturbations on the salient regions will be more likely to fool the deep models. Therefore, we introduce the saliency maps of video frames as prior knowledge for the generation of adversarial perturbations. As the perturbation are only generated for salient regions, it hence introduces spatial sparsity. To generate the saliency map for each frame, an efficient saliency detection approach proposed in \cite{hou2007saliency} and implement by OpenCV \cite{opencv_library} is applied. To control the area ratio of the salient region in the frame, we introduce a parameter $\varphi \in (0,1]$, and smaller $\varphi$ leads to smaller portion of salient regions. In the untarget attack, we initialize the mask $M$ to introduce spatial sparsity for all video frames based on $\varphi$. In the targeted attack, we also use the heuristic-based algorithm to obtain the descending order sequence of frames. After that, we only add spatial sparsity on the selected frame in $M$. In $M$, the pixels of salient regions is as $1$ while the rest is set $0$. We investigate different values of $\varphi$ in Section \ref{param_adjust}.

\begin{algorithm}
    \label{untar_algorithm}
    \SetKwInOut{Input}{Input}
    \SetKwInOut{Parameter}{Parameter}
    \SetKwInOut{Output}{Output}
    
    \Input{DNN $F$, clean video $x$, true label $y$, target class $y_{adv}$, an empty array $A$}
    \Output{Adversarial example $x_{adv}$.}
    \Parameter{$\omega$, $\varphi$, the number of update iterations $I$.}
    
    $M \gets \mathrm{SPATIAL}(x, \varphi)$\; 
    $M \gets \mathrm{Algorithm} \ \ref{key_frame}(F, x, y, y_{adv}, M, A, \omega)$\;
    $\theta = \frac{\hat{x} - x}{\Arrowvert \hat{x} - x \Arrowvert} $\;
    $\theta = \frac{\theta \times M}{\Arrowvert \theta \times M \Arrowvert}$\;
    \For{$t \gets 1$ \KwTo $I$}
    {$\hat{g} = \frac{g(\theta + \beta \textbf{u}) - g(\theta)}{\beta }\cdot \textbf{u}$\;
    $\theta = \theta - \eta \hat{g}$;
    }
    
    $x_{adv} = x + g(\theta) \times \theta$\;
    \Return $x_{adv}$
    \caption{Heuristic-based targeted attack algorithm.}
\end{algorithm}

\subsection{Overall Framework}
The whole process of the our method for the targeted attack is describe in Algorithm \ref{untar_algorithm}, where \textbf{SPATIAL} function performs a saliency detection with $\varphi$ for each frame, and initializes $M$ by setting the values of the salient regions to 1, the others to 0; \textbf{UPDATE} represents the zeroth order optimization to update $\theta$, . In the overall framework, we firstly initialize $M$ by \textbf{SPATIAL} to implement spatial sparsity, then combine the spatial sparsity with the temporal sparsity by Algorithm \ref{key_frame} to get the final $M$. After that, we obtain the temporal and spatial sparse $\theta$, and use \textbf{UPDATE} to update $\theta$ iteratively. Finally, the adversarial example can be found by the optimal direction.

\section{Experiments}
\label{experiment}
In this section, we test the performance of our heuristic based black-box attack algorithm with various parameters $\omega$ and $\varphi$ in reducing overall perturbations and the number of queries. Furthermore, we show a comprehension evaluation of our method on multiple video recognition models.

\subsection{Experimental Setting}
\subsubsection{Datasets.} We consider two widely used datasets for video recognition: UCF-101 \cite{su2009ucf} and HMDB-51 \cite{Kuehne11}. UCF-101 is an action recognition dataset that contains 13,320 videos with 101 action categories. HMDB-51 is a dataset for human motion recognition and contains a total of 7000 clips distributed in 51 action classes. Both datasets split 70\% of the videos as training set and the remaining 30\% as test set. During the evaluation, we use 16-frame snippets that uniform sampled from each video as input samples of target models.
\subsubsection{Metrics.} Four metrics are used to evaluate the performance of our method on various sides. 1) fooling rate (FR): the ratio of adversarial videos that are successfully misclassified. 
2) Median queries (MQ) \cite{ilyas2018black}: the median number of queries.
3) Mean absolute perturbation (MAP): denotes the mean perturbation of each pixel in the entire video. We use MAP$^{*}$ to denote the MAP of each pixel in the masked region. 
4) Sparsity (S): represents the proportion of pixels with no perturbations versus all pixels in a specific video. $S = 1 - \frac{1}{T}\sum_{t=1}^{T} \varphi_{t}$, where $\varphi$ is the area of the selected salient region at the corresponding frame, $\varphi_{t}$ is $\varphi$ of the t-th frame and T is the total number of frames.

\subsubsection{Threat Model}
We use the special case of the Partial-information setting \cite{ilyas2018black}. In this setting, the attacker only has access to the top-1 class $\widehat{y}$ and its probability $P(\widehat{y}|x)$, given a video $x$. Both untargeted and targeted black-box attacks are considered.
In the experiments, two video recognition models, Long-term Recurrent Convolutional Networks (LRCN) \cite{donahue2015long} and C3D \cite{hara2018can} are used as target models. LRCN \cite{donahue2015long} model uses Recurrent Neural Networks (RNNs) to encodes the temporal information and long-range dependencies on the features generated by CNNs. In our implementation, Inception V3 \cite{szegedy2016rethinking} is ultilzied to extract features for video frames and LSTM is utilized for video classification; C3D model \cite{hara2018can} applies 3D convolution to learn spatio-temporal features from videos with spatio-temporal filters for video classification. These models are the main methods of video classification. Table \ref{table_acc} summarizes the test accuracy of 16-frame snippets with these two models.

\begin{table}[t]
  \caption{Test Accuracy(\%) of the target models.}
  \label{table_acc}
  \centering
\begin{tabular}{ccc} \hline

Model        & UCF-101 & HMDB-51 \\ \hline\hline
C3D          & 85.88   & 59.57   \\
LRCN         & 64.92   & 37.42   \\ \hline
\end{tabular}
\end{table}

\subsection{Parameter Setting}
\label{param_adjust}
Note that we have two parameters need to set in Algorithm 2, one is the perturbations bound $\omega$ and the other one is the area ratio of salient regions $\varphi$. The parameter tuning is done on 30 videos that randomly sampled from the test set of UCF-101 and can be correctly classified by the target models. We do a grid search to find the most appropriate values for these two parameters. For $\omega$, we set it as $\{0, 3, 6, 9, 12, 15, \infty \}$ in the untargeted attack, as $\{0, 15, 30, 45, \infty\}$ in the targeted attack since the targeted attack has larger perturbations than the untargeted attack, and evaluate the attack performance on C3D model. Table \ref{tempo_table} shows the performance on the untargeted attack. Basically, large $\omega$ leads to sparse perturbations 
When $\omega$ is set to $\infty \ $, the sparsity value can be as low as 83.75\%, which means that only 2.6 ($16\times (1-83.75\%) \approx 2.6$) video frames will be selected to generate the adversarial perturbations. 
Therefore, to strikes a balance between the $\mathrm{MAP}^*$ and temporal sparsity, we set $\omega = 3$ in the untargeted attack to conduct subsequent experiments. Table \ref{tempo_targeted} lists the results on the targeted attack. The values of $\omega$ and $MQ$ is greater than those in the untargeted attack since the targeted attack has larger perturbations than the untargeted attack. We set $\omega = 30$ in experiments of the targeted attack.  

Similarly, we perform grid search to decide the value of $\varphi$. For the untargeted attack, we fix $\omega = 3$ and set parameter $\varphi$ as $\{0.2, 0.4, 0.6, 0.8, 1.0\}$ to evaluate the performance. Table \ref{spatio_table} lists the results. Note that, $\varphi = 1$ means that all the regions in the frame are selected. As can be seen, when the value of $\varphi$ is small, the median query number (MQ), mean absolute perturbations (MAP) will be reduced, while the mean absolute perturbations ($\mathrm{MAP}^*$) on the selected regions will be increased. When $\varphi = 0.2$, the sparsity value can be as high as 85.00\%, which results in a large $\mathrm{MAP}^*$ and lower fooling rate (FP). To balance the metrics $\mathrm{MAP}^*$, MQ and MAP, we set $\varphi$ as 0.6 for the untargeted attack. When $\varphi = 0.6$, it reduces the number of queries by 22.17\% and the overall perturbations (MAP) by 25.79\%. The results also suggest that the incursion of spatial sparsity brings to our method not just a reduction in the number of queries but a reduction in MAP. Table 4 lists the results for the targeted attack. Similarly, in order to balance the metrics $\mathrm{MAP}^*$, MQ and MAP, we set $\varphi$ as 0.8 for the targeted attack. 

\begin{table}[t]
\caption{Results of our algorithm with various $\omega$ in the untargeted attack. 
}
  \label{tempo_table}
  \centering
\begin{tabular}{l|lllll}
$\omega$  & \multicolumn{1}{l}{$FR(\%)$} & \multicolumn{1}{l}{$MQ$} & \multicolumn{1}{l}{$MAP$} & \multicolumn{1}{l}{$MAP^{*}$} & \multicolumn{1}{l}{$S(\%)$}\\ \hline
0  &    100 &   16085.0 &  3.7033 &  3.8449 & 17.69     \\ \hline
3  &    100 &   16085.0 &  3.6858 &  3.9667 & 25.19     \\ \hline
6  &    100 &   15996.0 &  3.7471 &  4.0328 & 23.94     \\ \hline
9  &    100 &   17527.0 &  3.7757 &  4.2862 & 34.19     \\ \hline
12 &    100 &   15912.5 &  3.8169 &  4.3646 & 36.44     \\ \hline
15 &    100 &   16795.0 &  3.7274 &  4.3429 & 36.69     \\ \hline
$\infty$ &100 &   14382.0 &  3.6039 &  7.9585 & 83.75     \\ 
\end{tabular}
\end{table}

\begin{table}[t]
\caption{Results of our algorithm with various $\varphi$ in the untargeted attack.}
  \label{spatio_table}
  \centering
\begin{tabular}{l|lllll}

$\varphi$  & \multicolumn{1}{l}{$FR(\%)$} & \multicolumn{1}{l}{$MQ$} & \multicolumn{1}{l}{$MAP$} & \multicolumn{1}{l}{$MAP^{*}$} & \multicolumn{1}{l}{$S(\%)$}\\ \hline
0.2  &    90  &   8770.0 &  1.5890 &  8.7153 & 85.00     \\ \hline
0.4  &    100 &   12336.0 &  2.6273 &  7.0203 & 68.84     \\ \hline
0.6  &    100 &   14125.0 &  3.2194 &  5.7604 & 54.25     \\ \hline
0.8  &    100 &   13845.0 &  3.4507 &  4.6347 & 40.33     \\ \hline
1.0  &    100 &   16085.0 &  3.6858 &  3.9667 & 25.19     \\ 
\end{tabular}
\end{table}

\begin{table}[t]
\caption{Results of our algorithm with various $\omega$ in the targeted attack. 
}
  \label{tempo_targeted}
  \centering
\begin{tabular}{l|lllll}
$\omega$  & \multicolumn{1}{l}{$FR(\%)$} & \multicolumn{1}{l}{$MQ$} & \multicolumn{1}{l}{$MAP$} & \multicolumn{1}{l}{$MAP^{*}$} & \multicolumn{1}{l}{$S(\%)$}\\ \hline
0  &    100 &   302230.50 &  9.7547 &  10.5442 & 8.54     \\ \hline
15  &    100 &   302230.50 &  9.7178 &  10.6463 & 11.67     \\ \hline
30  &    100 &   323615.50 &  8.5328 &  11.1309 & 26.88     \\ \hline
45  &    100 &   307470.00 &  10.6991 &  14.8790 & 35.00     \\ \hline
$\infty$ &    100 &   209826.00 &  5.0075 &  16.6886 & 71.98     \\ 
\end{tabular}
\end{table}

\begin{table}[t]
\caption{Results of our algorithm with various $\varphi$ in the targeted attack.}
  \label{spatio_targeted}
  \centering
\begin{tabular}{l|lllll}

$\varphi$  & \multicolumn{1}{l}{$FR(\%)$} & \multicolumn{1}{l}{$MQ$} & \multicolumn{1}{l}{$MAP$} & \multicolumn{1}{l}{$MAP^{*}$} & \multicolumn{1}{l}{$S(\%)$}\\ \hline
0.2  & 100 &  142253.5  & 11.7693  & 17.6957  &  44.17    \\ \hline
0.4  & 100 &  146720.0  & 13.4624  & 18.9002  & 36.54     \\ \hline
0.6  & 100 &  175194.5  & 11.4973  & 16.2451  & 34.58     \\ \hline
0.8  & 100 &  191216.0  & 10.7961  & 13.4766  & 22.58     \\ \hline
1.0  & 100 &  323615.0  & 8.5328  & 11.1309  & 26.88    \\ 
\end{tabular}
\end{table}

\begin{table*}[t]
    \caption{Untargeted and targeted attacks against C3D/LRCN Models. For all attack models, the Fooling Rate (FR) is 100\%. }
	\begin{center}\scalebox{0.80}{
		\begin{tabular}{ c|c|c|cccc||cccc } \hline
		    \multirow{2}{*}{Dataset} & \multirow{2}{*}{Target Model }& \multirow{2}{*}{Attack Model} & \multicolumn{4}{c||}{Untargeted attacks} & \multicolumn{4}{c}{Targeted attacks} \\ \cline{4-11}
		    &  &  & $MQ$ & $MAP$ & $MAP^{*}$ & $S(\%)$ & $MQ$ & $MAP$ & $MAP^{*}$ & $S(\%)$ \\ \hline \hline
		    \multirow{6}{*}{UCF-101} & \multirow{3}{*}{C3D} &  Opt-attack \cite{cheng2018query}               & 17997.5 & 4.2540 & \textbf{4.2540} & 0.00 & 207944.5 & 9.0906 & \textbf{9.0906} & 0.00 \\ 
		  &  & Our (Temp.)     & 16292.0 & 4.0895 & 4.3642  & 21.19  & 313229.0 & \textbf{7.8069} & 10.4700 & 28.00 \\ 
            & &Our (Temp. + Spat.)   & \textbf{12940.0} & \textbf{3.0346}  & 5.5189 & \textbf{54.33}  & \textbf{167217.0} & 10.8588 & 15.4904 & \textbf{34.28}   \\ \cline{2-11}
            & \multirow{3}{*}{LRCN} & Opt-attack \cite{cheng2018query}     & 12359.5 & \textbf{1.8320} & \textbf{1.8320} & 0.00 & 445279.0 & 13.4795 & \textbf{13.4795} & 0.00\\ 
            &  & Our (Temp.)             & 14713.5 & 1.8754 & 1.8794  & 17.19    & 566719.0 & 11.7858 & 14.7894 & 23.33  \\
            &  & Our (Temp. + Spat.)   & \textbf{8421.5} & 1.8383 & 3.0848 & \textbf{47.50}  & \textbf{399655.0} & \textbf{11.2066} & 19.8620 & \textbf{46.92}\\ \hline \hline
		    \multirow{6}{*}{HMDB-51} & \multirow{3}{*}{C3D} &  Opt-attack \cite{cheng2018query}               & 14509.5 & 2.8930 & \textbf{2.8930} & 0.00  & 205286.5 & \textbf{6.5704} & \textbf{6.5704} & 0.00 \\ 
		  &  & Our (Temp.)     & 13536.5 & 2.9214 & 3.2010  & 26.94  & 196371.5 & 8.3599 & 10.6761 & 21.88 \\ 
            & &Our (Temp. + Spat.)   & \textbf{10616.0} & \textbf{2.3765}  & 4.4574 & \textbf{57.04}  & \textbf{144917.5} & 9.6109 & 12.2993 & \textbf{28.70}   \\ \cline{2-11}
            & \multirow{3}{*}{LRCN} & Opt-attack \cite{cheng2018query}     & 18655.0 & 2.7586 & \textbf{2.7586} & 0.00 & 224414.0 & \textbf{3.8598} & \textbf{3.8598} & 0.00\\ 
            &  & Our (Temp.)             & 15369.5 & 2.8011 & 2.8923  & 24.22    & 339367.0 & 4.0618 & 5.5601 & 28.75  \\
            &  & Our (Temp. + Spat.)   & \textbf{13311.5} & \textbf{1.5390} & 2.8302 & \textbf{62.03}  & \textbf{206120.0} & 12.7966 & 18.1835 & \textbf{42.87}\\ \hline \hline
		    \end{tabular}}
	\end{center}
	\label{tab:compare_baseline}
\end{table*}

\begin{figure*}[h]
  \centering
  \includegraphics[scale=0.7]{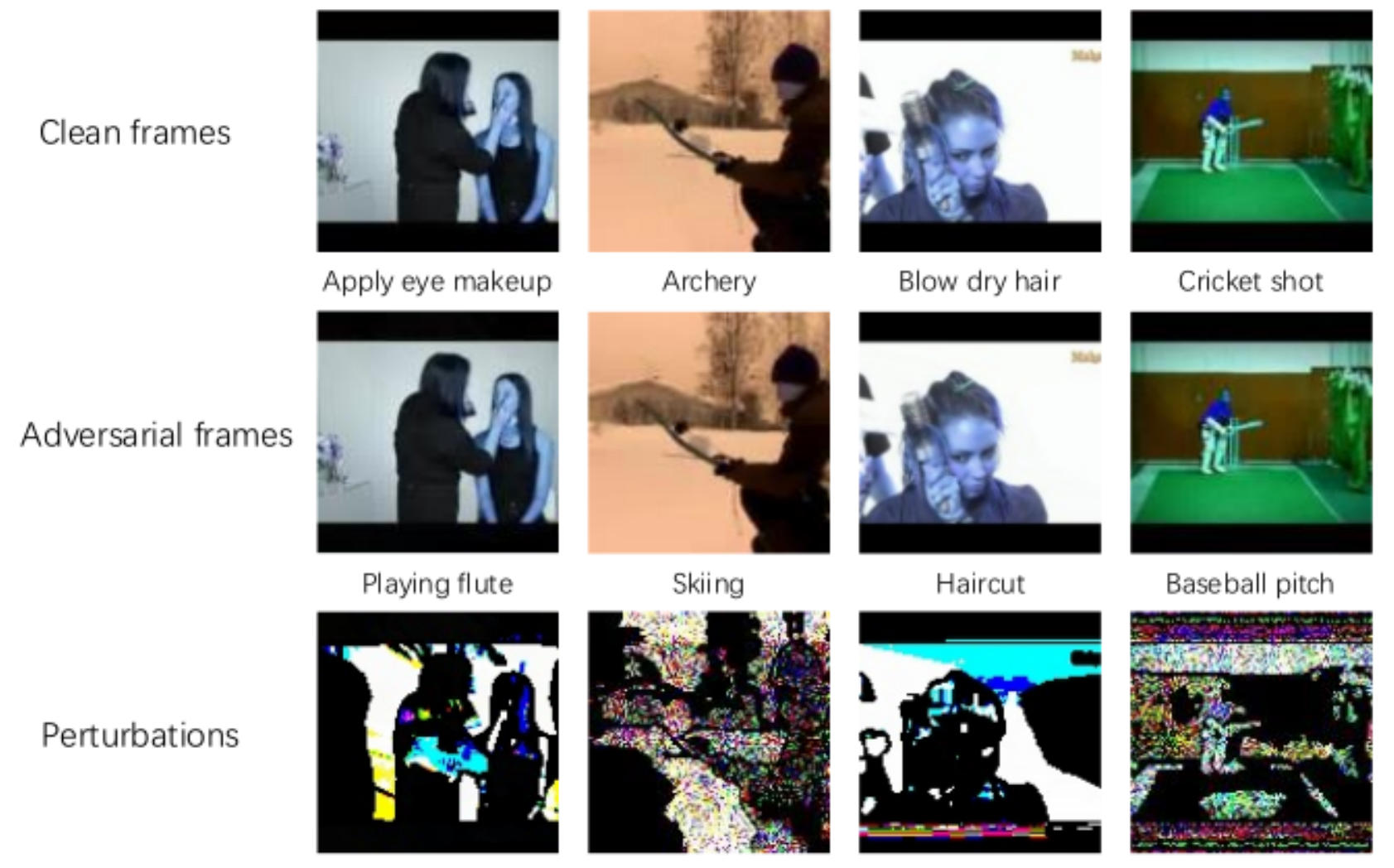}
  \caption{Examples of adversarial frames generated in the untargeted attack by our method. The clean frames are show in the top row; the corresponding adversarial frames are in the middle row and the perturbations are show in the bottom row. The perturbations are visualized by re-scaling them into the range of 0-255.}
  \label{fig:overview}
\end{figure*}

\subsection{Performance Comparison}
\label{attack_res}
We compare our method with the Opt-attack \cite{cheng2018query} which originally proposed to attack image classification models under black-box setting. We directly extend Opt-attack to attack video models for comparison. Besides, we also compare with one variant of our method by removing the spatial sparsity module. In this setting, only temporal sparsity is considered. The evaluations are performed with two video recognition models on two datasets. 

Table \ref{tab:compare_baseline} lists the performance comparisons regarding to the untargeted attack as well the targeted attack on UCF-101 dataset and HMDB-51 dataset. For untargeted attack, we have the following observations. First, compared to Opt-attack, our method that considers both temporal and spatial sparsity significantly reduces the number of queries. For LRCN and C3D model, the number of queries has been reduced by more than 28\% on both datasets. Second, compared to introducing temporal sparsity, introducing spatial sparsity is more effective in reducing the query numbers. For example, for C3D, introducing temporal sparsity alone helps to reduce the query numbers around 9\% while introducing both spatial and temporal sparsity reduce the query numbers for more than 28\% on UCF-101. By looking into the results, we found that around 12\% of the queries are spent on searching a set of key frames that maximize the sparsity value while keeping the mean absolute perturbations (MAP) lower than the given bound. Third, on most cases, introducing temporal and spatial sparsity helps to make the adversarial perturbations much sparser and hence reduce the mean absolute perturbations (MAP). In summary, introducing temporal and spatial sparsity increases the query efficiency and helps to achieve human-imperceptible perturbations.

For the targeted attack, similar trends are observed as the untargeted attack. By introducing both spatial and temporal sparsity, our method significantly reduces the query numbers. On UCF-101 dataset, the query numbers can be reduced from 445,279 to 399,655 for LRCN model. For C3D model, the query numbers have been reduced by more than 19.59\% on both datasets.
Although spatial sparsity significantly reduces the number of queries, it increases MQ, MAP and MAP$^{*}$ a large margin on the targeted attack. Compared to the untargeted attack, MQ, MAP as well MAP$^{*}$ on the targeted attack are much higher. One major reason is that compared to the untargeted attack, the target attack are much more difficult. Besides, to make sure that it achieves 100\% of the fooling rate, the MAP as well MAP$^{*}$ increased significantly when introducing the spatial sparsity. The results basically suggest that for the targeted attack, the selected saliency regions from the clean video frames may not contribute much to the classification of targeted classes, hence it increases the perturbations in order to successfully fool the recognition model, the generated sparse directions cause the targeted attack to fall into local optimal solutions and stop the direction updating early. In addition, introducing temporal sparsity increases the query numbers because  39.75\% of the queries are spent on searching a small set of keyframes in order to achieve temporal sparsity.

Figure 2 further shows four examples of adversarial frames generated by the proposed method. For all the examples, our method successfully fools the recognition model. For example, in the first example, the ground-truth label for the video is ``Apply eye makeup", by adding the generated human-imperceptible adversarial perturbations (second row), the model tends to predict a wrong label ``Playing flute" at the top-1 place. By re-scaling the adversarial perturbations to the range of 0-255, we visualize the adversarial perturbations in the third row. As can be seen, the generated adversarial perturbations are basically quite sparse and most of them are focused on the foreground of the key frames.

\section{Conclusion}
In this paper, we proposed a heuristic black-box adversarial attack algorithm for video recognition models. To reduce query numbers and improve attack efficiency, our method explores the sparsity of adversarial perturbations in both temporal and spatial domains. 
Our algorithm is adaptive to multiple target models and video datasets and enjoys global sparsity and query efficiency improvement. Moreover, the experimental results demonstrate that video recognition models are vulnerable to adversarial attack, and our algorithm achieves small human-imperceptible perturbation using fewer queries. The most pertinent area of future work is to further investigate the black-box attack for the targeted attack using fewer queries.

\section{Acknowledgments}
The work was funded by the National Research Foundation, Prime Ministers Office, Singapore under its IRC@Singapore Funding Initiative, and the NSFC Projects (No.61806109). The work was also funded by the Jilin Provincial Key Laboratory of Big Data Intelligent Computing (20180622002JC), the Education Department of Jilin Province (JJKH20180145KJ), and the startup grant of the Jilin University, the Bioknow MedAI Institute (BMCPP-2018-001), the High Performance Computing Center of Jilin University, and the Fundamental Research Funds for the Central Universities, JLU.

\bibliographystyle{aaai.bst}
\bibliography{ref.bib}
\end{document}